\documentclass[11pt,a4paper]{article}
\usepackage[hyperref]{acl2019}

\usepackage{natbib}
\usepackage{hyperref}
\usepackage{times,latexsym}
\usepackage{url}
\usepackage[T1]{fontenc}
\usepackage[utf8x]{inputenc}
\aclfinalcopy 
\def\aclpaperid{1609} 

\usepackage{graphicx}
\usepackage{booktabs}
\usepackage{multirow}
\usepackage{amsmath}
\usepackage{appendix}
\usepackage{todonotes}
\usepackage{breqn}

\title{Symbolic inductive bias for visually grounded learning of spoken language}
\author{    Grzegorz Chrupała\\
      Tilburg University \\
      {\tt g.chrupala@uvt.nl }}
\date{}
\begin{document}
\maketitle

\begin{abstract}
\noindent A widespread approach to processing spoken language is to first automatically transcribe it into text. An alternative is to use an end-to-end approach: recent works have proposed to learn semantic embeddings of spoken language from images with spoken captions, without an intermediate transcription step. We propose to use multitask learning to exploit existing transcribed speech within the end-to-end setting.
We describe a three-task architecture which combines the objectives of matching spoken captions with corresponding images, speech with text, and text with images. We show that the addition of the {\sc speech/text} task leads to substantial performance improvements on image retrieval when compared to training the {\sc speech/image} task in isolation. We conjecture that this is due to a strong inductive bias transcribed speech provides to the model, and offer supporting evidence for this. 
\end{abstract}

\section{Introduction}
Understanding spoken language is one of the key capabilities of intelligent systems which need to interact with humans. 
Applications include personal assistants, search engines, vehicle navigation systems and many others. 
The standard approach to understanding spoken language both in industry and in research has been to  decompose the problem into two components arranged in a pipeline: Automatic Speech Recognition (ASR) and  Natural Language Understanding (NLU). 
The audio signal representing a spoken utterance is first transcribed into written text, which is subsequently processed to extract some semantic representation of the utterance. 
Recent works have proposed to learn semantic embeddings of spoken language by using photographic images of everyday situations matched with their spoken captions, without an intermediate transcription step \citep{harwath2016unsupervised,chrupala2017representations}. The weak and noisy supervision in these approaches is closer to how humans learn to understand speech by grounding it in perception and thus more useful as a cognitive model. It can also have some practical advantages: in certain circumstances it may be easier to find or collect speech associated with images rather than transcribed speech -- for example when dealing with language whose speakers are illiterate, or for languages with no standard writing system (note that even some languages with many millions of speakers, like Cantonese, may not have a standardized writing system). 
On the other hand, the learning problem in this type of framework is less constrained, and harder, than standard ASR. 

In order to alleviate this shortcoming, we propose to use multitask learning (MTL) and exploit transcribed speech within the end-to-end visually-grounded setting, and thus combine some features of both the pipeline and end-to-end approaches. Incorporating speech transcriptions into the end-to-end architecture via multi-task learning measn that the amount of transcribed speech and its quality do not need to be as high as needed for training an ASR system within the pipeline architecture, since the role of this data is only to guide the end-to-end model via an auxiliary task. 

We describe a three-task architecture which combines the main objective of matching speech with images with two auxiliary objectives: matching speech with text, and matching text with images.

The plain end-to-end {\sc speech/image} matching task, modeled via standard architectures such as recurrent neural networks, lacks a language-specific learning bias. This type of model may discover in the course of learning that speech can be represented as a sequence of symbols (such as for example phonemes or graphemes), but it is in no way predisposed to make this discovery. Human learners may be more efficient at least in part thanks to their innate inductive bias  whereby they assume that language is symbolic. They arguably acquired such bias via the process of evolution by natural selection. In the context of machine learning, inductive bias can instead be injected via multi-task learning, where supervision from the secondary task guides the model towards appropriately biased representations. 

Specifically, our motivation for the {\sc speech/text} task is to encourage the model to learn speech representations which are correlated with the encoding of spoken language as a sequence of characters. 
Additionally, and for completeness, we also consider a second auxiliary task matching text to images.

Our contribution consists in formulating and answering the following questions: 
\begin{itemize}
    \item Do the auxiliary tasks improve the main {\sc speech/image} task? The {\sc speech/text} task helps but we have no evidence of the {\sc text/image} task improving performance.
    \item If so, is this mainly because MTL allows us to exploit extra data, or because the additional task injects an appropriate inductive bias into the model?
    The inductive bias is key to the performance gains of MTL, while extra data makes no impact.
    \item Which parameters should be shared between tasks and which should be task specific?
    Best performance is achieved by sharing only the lower layers of the speech encoder.
    \item What are the specific effects of the symbolic inductive bias on the learned representations? {\sc speech/text} contributes to make the encoded speech more speaker invariant, and more strongly correlated to the written or phonetically represented form of the utterances. 
\end{itemize}


\section{Related work}

\subsection{Visually grounded semantic embeddings of spoken language}

The most relevant strand of related work is on visually-grounded learning of (spoken) language. It dates back at least to \citet{Roy2002113}, but has recently attracted further interest due to better-performing modeling tools based on neural networks. 

\citet{harwath2015deep} collect spoken descriptions for the Flick8K captioned image dataset and present a model which is able to map pre-segmented spoken words to aspects of visual context. \citet{harwath2016unsupervised} describe a larger dataset of images paired with spoken captions (Places Audio Caption Corpus) and present an architecture that learns to project images and unsegmented spoken captions to the same embedding space. The sentence representation is obtained by feeding the spectrogram to a convolutional network. Further elaborations on this setting include \citet{P17-1047}, which shows a clustering-based method to identify grounded words in the speech-image pairs, and \citet{harwath2018jointly} which constructs a three-dimensional tensor encoding affinities between image regions and speech segments.

The work of \citet{chrupala2017representations} is similar in that it exploits datasets of images with spoken captions, but their grounded speech model is based around multi-layer Recurrent Highway Networks, and focuses on quantitative analyses of the learned representations. They show
that the encoding of meaning tends to become richer in higher layers, whereas encoding of form tends
to initially increase and then stay constant or decrease. \citet{K17-1037} further analyze the representations of the same model and show that phonological form is reliably encoded in the lower recurrent layers of the network but becomes substantially attenuated in the higher layers.

\citet{drexler2017analysis} also analyze the representations of a visually grounded speech model with view of using such representations for unsupervised speech recognition, and show that they contain more linguistic and less speaker information than filterbank features.

\citet{kamper2017visually} use images as a pivot to learn to associate textual labels with spoken utterances, by mapping utterances and images into joint semantic space. After labeling the images with an object classifier, these labels can be further associated with utterances, providing bag-of-words representation of spoken language which can be useful in speech retrieval.

\subsection{Multi-task learning for speech and language}
The concept of multi-task learning (MTL) was introduced by \citet{caruana1997multitask}. Neural architectures widely used in the fields of speech and language processing make it easy to define parameter-sharing architectures and exploit MTL, and thus there has been a recent spurt of reports on its impact.

Within Natural Language Processing (NLP), \citet{44928} explore sharing encoders and decoders in a sequence-to-sequence architecture for translation, syntactic parsing, and image captioning, and show gains on some configurations. \citet{E17-2026} investigate which particular pairs of NLP tasks lead to gains, concluding that learning curves and label entropy of the tasks may be used as predictors. \citet{mccann2018natural} propose a 10-task NLP challenge, and a single MTL model which performs reasonably well on all tasks.

\citet{P16-2038} show that which parameters are shared in a multi-task architecture matters a lot: they find that when sharing parameters between syntactic chunking or supertagging and POS tagging as an auxiliary task, it was consistently better to only share the lower-layers of the model. 
Relatedly, \citet{D17-1206} propose a method of training NLP tasks at multiple levels of complexity by growing the depth of the model to solve increasingly more difficult tasks.
\citet{Swayamdipta2018SyntacticSF} use similar ideas and show that 
syntactic information can be incorporated in a semantic task with MTL, using auxiliary syntactic tasks without building full-fledged syntactic structure at prediction time.

MTL can lead to a bewildering number of choices regarding which tasks to combine, which parameters to share and how to schedule and weight the tasks. Some recent works have suggested specific approaches to deal with this complexity: \citet{ruder2017learning} propose to learn from data which parameters to share in MTL with sluice networks and show some gains on NLP tasks.
\citet{kiperwasser2018scheduled} investigate how to interleave learning syntax and translation and how to schedule these tasks.


Several works show that exploiting MTL via the use of multiple language versions of the same or comparable data leads to performance gains \citep[e.g.][]{lee2017fully,Q17-1024,de2018parameter}. 
\citet{gella2017image} and \citet{kadar2018lessons} learn visual semantic embeddings from textual-visual datasets and show gains from additional languages which reuse the same encoder. \citet{kadar2018lessons} additionally show that an extra objective linking the languages directly rather than only via the visual modality provides additional performance gains.
In the context of audio-visual data, \citet{harwath2018vision} applies a type of MTL in the setting where there are images paired with descriptions in English and Hindi. They project the images, English speech and Hindi speech into a joint semantic space, and show that training on multiple tasks matching  both languages to images works better compared to only using a single monolingual task.

MTL has also recently seen some success in speech processing. 
Similar to what we see in machine translation, in ASR parameter sharing between different languages is also beneficial \citep{6639348}. More recently, \citet{DBLP:journals/corr/abs-1802-07420} show that exploiting this effect is especially useful for low-resource languages.

\citet{6639012} apply MTL for phone recognition with three lower-level auxiliary tasks and show noticeable reductions in error rates.
\citet{DBLP:conf/interspeech/ToshniwalTLL17} use MTL for conversational speech recognition with lower-level tasks (e.g.\ phoneme recognition) in an encoder-decoder model for direct character transcription. 
\citet{7953071} learn to align utterances with phonetic transcriptions in a lower layer and graphemic transcriptions in the final layer, exploiting again the relation between task level of complexity and levels of neural architecture in a MTL setting. They also show a benefit of sharing model parameters between different varieties of the same language, specifically US, British, Indian and Australian English.
\citet{DBLP:journals/corr/abs-1710-08377} demonstrate the effectiveness of transfer from generic audio classification to speech command recognition, which can also be considered a particular instance of MTL.

\paragraph{How our work fits in.}
The current paper uses an intuition also present in several of the works mentioned above: namely that an end-to-end model which needs to induce several levels of intermediate latent representations should be guided to find useful ones by including auxiliary prediction tasks at the intermediate layers. These auxiliary prediction tasks typically use lower-level linguistically-motivated structures such as phonemes for end-to-end ASR, or syntactic trees for semantic parsing. 

The present study extends this setting to a full speech-to-semantics setup: the main task is to take spoken language as input and learn a semantic representation based on feedback from the visual modality, while an ASR-like task ({\sc speech/text matching}) is merely auxiliary. The lower-level linguistic structures in our case are the sequences of phoneme-like units approximated by the written form of the language.

\section{Methods}

\subsection{Models}
\begin{figure}[ht]
    \centering
    \includegraphics[scale=0.6]{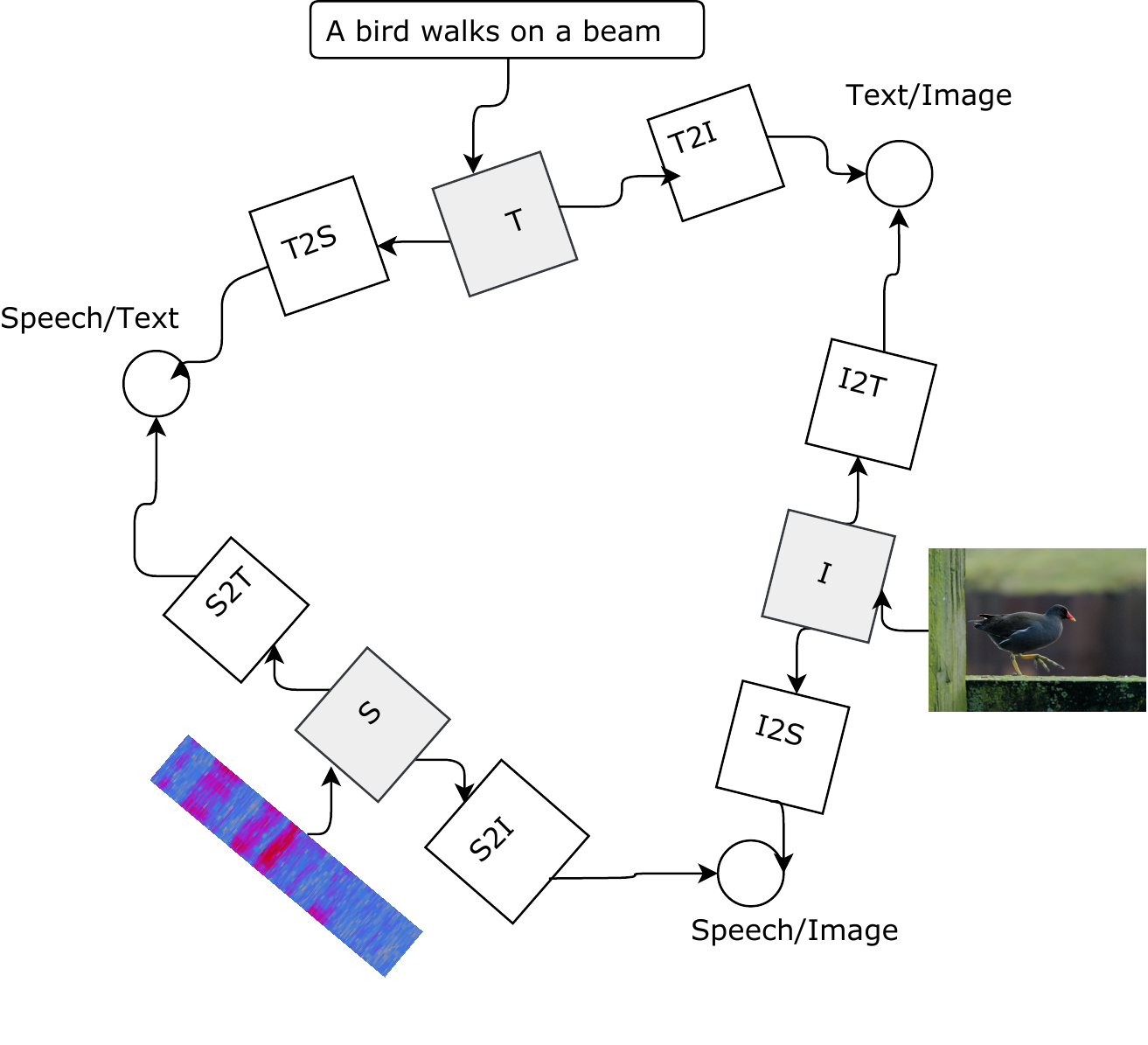}
    \caption{Overview of the task architecture. {\sc t}: shared text encoder, {\sc s}: shared speech encoder, {\sc i}: shared image encoder. The notation {\sc x2y} stands for an encoder for input type {\sc x} which is only used for the loss between encoded input types {\sc x} and {\sc y}.}
    \label{fig:core-model}
\end{figure}

The modeling framework uses a multi-task setup. The core model is a three-task architecture depicted in Figure~\ref{fig:core-model}: there are three encoders, one for each modality: speech, image, and text. Each modality has a shared encoder which works directly on the input modality, and two specialized encoders which take as input the encoded data from the shared encoder. The three tasks correspond to three losses (depicted with circles in the figure): each loss works with a pair of modalities and attempts to minimize the distance between matching encoded items, while maximizing the distance between mismatching ones. For a pair of modalities with encoded objects $\mathbf{u}$ and $\mathbf{i}$, the loss is defined as follows

\begin{dmath}
  \sum_{\mathbf{u},\mathbf{i}} \left(\sum_{\mathbf{u}'} \max[0, \alpha + d(\mathbf{u},\mathbf{i}) - d(\mathbf{u}',\mathbf{i})] +
    \sum_{\mathbf{i}'} \max[0, \alpha + d(\mathbf{u},\mathbf{i}) - d(\mathbf{u},\mathbf{i}')] \right)
\end{dmath}
where $(\mathbf{u,i})$ are matching objects (for example an utterance and a matching image), and $(\mathbf{u',i})$ and $(\mathbf{u,i'})$ are mismatched objects within a batch, while $d(\cdot, \cdot)$ is the cosine distance between encoded objects.

The {\sc speech/image} part of the architecture is based on the grounded speech model from \citet{chrupala2017representations}, with the main difference being that these authors used Recurrent Highway Networks \citep{pmlr-v70-zilly17a} for the recurrent layers, while we chose the simpler Gated Recurrent Unit networks \citep{chung2014empirical}, because they have optimized low-level CUDA support which makes them much faster to run and enables us to carry out an at least somewhat comprehensive set of experiments.

\subsection{Image Encoders}
The shared image encoder {\sc i} is a pretrained, fixed Convolutional Neural Network which outputs a vector with  image features; specifically, the activations of the pre-classification layer. The modality-specific encoders {\sc i2s} and {\sc i2t} are linear mappings which take the output of {\sc i}.
\subsection{Speech Encoders}
\label{sec:encoder}
The shared encoder {\sc s} consists of a 1-dimensional convolutional layer which subsamples the input, followed by a stack of recurrent layers. The modality specific encoders {\sc s2t} and {\sc s2i} consist of a stack of recurrent layers, followed by an attention operator. 
The encoder $\mathrm{S}$ is defined as follows:
\begin{equation}
  \label{eq:encode_s}
  \mathrm{S}(\mathbf{x}) = \mathrm{GRU}_{\ell} (\mathrm{Conv}_{s,d,z}(\mathbf{x}))
\end{equation}
where  $\mathrm{Conv}$ is a convolutional layer with kernel size
$s$, $d$ channels, and stride $z$, and $\mathrm{GRU}_\ell$ is a stack of $\ell$ GRU layers. 
An encoder of modality  {\sc x} is defined as 
\begin{equation}
    \label{eq:encode_s2x}
    \mathrm{S2X}(\mathbf{x}) = \mathrm{unit}(\mathrm{Attn}(\mathrm{GRU}_{\ell} (\mathbf{x})))
\end{equation}
where $\mathrm{Attn}$ is the attention operator and $\mathrm{unit}$ is L2-normalization. Note that for the case $\ell=0$ $\mathrm{GRU}_{\ell}$ is simply the identity function. The attention operator computes a weighted sum of the RNN activations at all timesteps:
\begin{equation}
  \label{eq:attn}
  \mathrm{Attn}(\mathbf{x}) = \sum_t \alpha_t \mathbf{x}_t
\end{equation}
where the weights $\alpha_t$ are determined by an MLP with learned parameters
$\mathbf{U}$ and $\mathbf{W}$, and passed through the timewise softmax function:
\begin{equation}
  \label{eq:alpha_t}
  \alpha_t = \frac{\exp(\mathbf{U}\tanh(\mathbf{Wx}_t))}{\sum_{t'} \exp(\mathbf{U}\tanh(\mathbf{Wx}_{t'}))}
\end{equation}

\subsection{Text Encoders}
The text encoders are defined in the same way as the speech encoders, with the only difference being that the convolutional layer is replaced by an embedding layer, i.e.\ a lookup table mapping characters to embedding vectors.

\subsection{Multi-tasking}
The model is trained by alternating between the tasks, and updating the parameters of each task in turn. Note that the input data for the three tasks can be the same, but can also be partly or completely disjoint. We 
report two conditions
\begin{itemize}
    \item {\sc Aligned}: all tasks use the same parallel data;
    \item {\sc Non-aligned}: the data for the {\sc Speech/Text} task is disjoint from the data for the other two tasks.
\end{itemize}

We consider the {\sc non-aligned} condition somewhat more realistic, in that it is easier to find separate datasets for each pair of modalities than it is to to find a single dataset with all three modalities. However the main reason to including both conditions is that it allows us to disentangle via which mechanism MTL contributes: by enabling the use of extra data, or by enforcing an inductive bias.
\subsection{Architecture variants}
There is a multitude of ways in which the details of the core architecture can be varied. in order to reduce them to a manageable number we made the following choices:
\begin{itemize}
    \item Keep the image encoder simple and fixed. 
\item Keep the architecture of the encoders fixed, and only vary encoder depth and the degree of sharing.
\end{itemize}

In addition to variants of the full three-task model, we also have single-task and two-task baselines which are the three-task model with the {\sc speech/text} and {\sc text/image} tasks completely ablated, or with only the {\sc text/image} task ablated. Note that we do not include a condition with only the {\sc speech/text} task ablated, as the two remaining tasks do not share any learnable parameters (since {\sc I} is fixed).

\subsection{Evaluation metrics}
Below we introduce metrics evaluating performance on the image retrieval task, as well as additional analytical metrics which quantify some aspects of the internal representation learned by the encoders.

\paragraph{Evaluating image retrieval}
In order to evaluate how well the main {\sc speech/image} task performs we report the recall at 10 (R@10) and median rank (Medr) for the {\sc speech/image} task: utterances in the development set are encoded via {\sc s2i} and images via {\sc i2s}. For each utterance the images are ranked in order of cosine distance; R@10 counts the mean proportion of correct images among top 10 ranked images, while Medr gives the median of the ranks of the correct image (where correct image counts as image originally paired with the utterance).

\paragraph{Invariance to speaker}
We measure how invariant the utterance encoding is to the identity of the speaker; in principle it is expected and desirable that the utterance encoding captures the meaning of the spoken language rather than other aspects of it such as who spoke it. To quantify this invariance we report the accuracy of an L2-penalized logistic regression model on the task of decoding the identity of the speaker from the output of the {\sc s2i} encoder. The logistic model is trained on $\frac{2}{3}$ of the development data and tested on the remaining $\frac{1}{3}$.

\paragraph{Representational similarity}
Representational Similarity Analysis \citep{kriegeskorte2008representational} gauges the  correlation between two sets of pairwise similarity measurements. Here we use it to quantify the correlation of the learned representation space with the written text space and with the image space.
For the encoder representations, the pairwise similarities between utterances are given by the cosine similarities. For the written form, the similarities are the inverse of the normalized Levenshtein distance between the character sequences encoding each pair of utterances:
\begin{equation}
\mathrm{sim_{text}}(a, b) = 1 - \frac{D(a, b)}{\max(|a|, |b|)}
\end{equation}
where $D(a, b)$ is the Levenshtein distance and $|\cdot|$ is string length.
We compute the Pearson correlation coefficient between two similarity matrices on the upper triangulars of the each matrix, excluding the diagonal.

\paragraph{Phoneme decoding}
A direct way of measuring whether neural representations of speech are biased towards encoding symbols is to try to decode the phonemes from the activation patterns aligned with a phonetic transcription of the utterance. We follow the methodology of \citet{alishahi2017encoding} and train an L2-penalized logistic regression model on the output of the {\sc s} encoders for phonemes from 2,500 utterances and report classification accuracies on data from 2,500 heldout utterances.

\subsection{Experimental settings}

\paragraph{Data}
The {\sc speech/image} and {\sc text/image} tasks are always trained on the Flickr8K Audio Caption Corpus \citep{harwath2016unsupervised}, which is based on the original Flickr8K dataset \citep{hodosh2013framing}. Flickr8K consists of 8,000 photographic images depicting everyday situations. Each image is accompanied by five brief English descriptions produced by crowd workers. Flickr8K Audio Caption Corpus enriches this data with spoken versions of these descriptions, read aloud and recorded by crowd workers. The total amount of speech in this dataset is approximately 34 hours. 
One thousand images are held out for validation, and another one thousand for the  test set, using the splits provided by \citet{karpathy2015deep}. 
In the {\sc aligned} condition the {\sc speech/text} task is also trained on this data. In the {\sc non-aligned} condition, we train the {\sc speech/text} task on the Libri dataset \citep{7178964} which consists of approximately 1,000 hours of read English speech, derived from read audiobooks. There are 291,630 sentences in the corpus, of which 1,000 are held out for validation. 

We preprocess the audio by extracting 12-dimensional mel-frequency cepstral
coefficients (MFCC) plus log of the total energy. We use 25 millisecond windows, sampled every 10 milliseconds. The shared image encoder is fixed and consists of 4096 dimensional activations of the pre-classification layer of VGG-16 \citep{simonyan2014very} pre-trained on Imagenet \citep{Russakovsky2015}.
\paragraph{Hyperparameters}
Most of the hyperparameters are based from especially \citet{chrupala2017representations}. The models are trained for a maximum of 25 epochs with Adam, with learning rate 0.0002, and gradient clipping at 2.0. The loss function's margin parameter is $\alpha=0.2$. The GRUs have 1024 dimensions. The convolutional layer has 64 channels, kernel size of 6 and stride 2. The hidden layer of the attention MLP is 128. The linear mappings {\sc i2s} and {\sc i2t} project 4096 dimensions down to 1024. We apply early stopping and pick the results of each run after the epoch for which it scored best on R@10. We run three random initializations of each configuration.
\paragraph{Multi-task training}
We use a simple round-robin training scheme: we alternate between tasks, and for each task update the parameters of that task as well as the shared parameters based on supervision from one batch of data. The data ordering for each task is independent, both in the {\sc aligned} and {\sc non-aligned} condition: for each epoch we reshuffle the dataset associated to each task and iterate through the batches until the smallest dataset runs out. This procedure makes sure that the only difference between the {\sc aligned} and {\sc non-aligned} conditions is the actual data and not other aspects of training.

\paragraph{Repository}
The code needed to reproduce our results and analyses is available at \href{https://github.com/gchrupala/symbolic-bias}{https://github.com/gchrupala/symbolic-bias}.

\section{Results}

\begin{table*}[!htb]
    \centering
    \begin{tabular}{lll rrr rrr rr}
  & Data & Tasks & \sc s & \sc t & \sc s2i & \sc s2t &\sc  t2s & \sc t2i & R@10 & Medr \\\toprule
1 &    NA   &  1 & 2 & . & 2   & .   & .   & .   & 0.218 &  63.8 \\ 
2 &   Aligned & 2 & 2 & 1 & 2   & 0   & 0   & .   & 0.279 &  42.3 \\
3 & Non-aligned & 2 & 2 & 1 & 2   & 0   & 0   & .   &  0.280 &  41.3 \\
\midrule 
4 & \multirow{5}{*}{ Aligned } & \multirow{5}{*}{3} & 2 & 1 & 1 & 0 & 0 & 1 & 0.280 &  43.0 \\
5 & & & 2 & 1 & 1 & 1 & 1 & 1 & 0.266 &  44.3 \\
6 & & & 2 & 1 & 2 & 0 & 0 & 1 & \bf 0.281 &  \bf 39.7 \\
7 & & & 2 & 1 & 2 & 1 & 1 & 1 & 0.270 &  44.3 \\
8 & & & 4 & 1 & 0 & 0 & 0 & 0 & 0.255 & 48.3 \\\midrule
9 & \multirow{5}{*}{ Non-aligned } & \multirow{5}{*}{3} & 2 & 1 & 1 & 0 & 0 & 1 &   0.275 &  42.8 \\
10 & & & 2 & 1 & 1 & 1 & 1 & 1 &  0.257 &  49.8 \\
11 & & & 2 & 1 & 2 & 0 & 0 & 1 & 0.280 &  41.7 \\
12 & & & 2 & 1 & 2 & 1 & 1 & 1 &  0.252 &  50.7 \\
13 & & & 4 & 1 & 0 & 0 & 0 & 0 &  0.223 &  59.3 \\

    \end{tabular}
    \caption{Results on the validation set with varying model configuration. R@10 is recall at 10 for the Speech/Image task, Medr is the median rank for the same task.  All scores are averages over $3$ runs with different random initializations; models were run for 25 epochs with early stopping with R@10 as a criterion. The numbers (1, 2) in the columns corresponding to encoders specify the number of RNN layers in each encoder; zero (0) indicates the encoder only consists of the self-attention with no RNN layers; dot (.) indicates the whole task in which the encoder participates is ablated.}
    \label{tab:core-results}
\end{table*}

\begin{table*}[!htb]
    \centering
    \begin{tabular}{ll rrr rrr rr}
Data & Tasks & \sc s & \sc t & \sc s2i & \sc s2t &\sc  t2s & \sc t2i & R@10 & Medr \\\toprule
NA & 1 & \multicolumn{6}{c}{ \citet{harwath2015deep} }             & 0.179 & - \\
NA & 1 & \multicolumn{6}{c}{ \citet{chrupala2017representations} } & 0.253 & 48 \\
\midrule
NA & 1 & 2 & . & 2   & .   & .   & .   &  0.244 &  51  \\ 
Aligned & 3 & 2 & 1 & 2 & 0 & 0 & 1         &  0.296  &  34 \\
\end{tabular}    
\caption{Results on the test set, obtained by using the best run/epoch determined on the validation data. The first two rows show the numbers reported in previous work.}
\label{tab:core-results-test}
\end{table*}    

Table~\ref{tab:core-results} shows the evaluation results on the validation data, on the image retrieval task of 13 configurations of the model, including three versions with one or two tasks ablated. 

Table~\ref{tab:core-results-test} shows the results on the test set with the 1-task baseline model and the best performing configuration compared to previously reported results on this dataset. 
As can be seen the baseline model is a bit worse than the best reported result on this data, while the 3-task model is much better.

\section{Discussion}
Below we discuss and interpret the patterns in performance on image retrieval as measured by Recall@10 and median rank.

\paragraph{Impact of tasks}
The most striking result is the large gap in performance between the 1-task condition (row 1) and most of the other rows. Comparing row 1 versus rows 2 and 3 we see that adding the {\sc speech/text} task leads to a substantial improvement.  However,  comparing rows 2 and 3 versus rows 6 and 11,  it seems that the addition of {\sc text/image} task does not seem to have a major impact on performance, at least to the extent that can be gleaned from the experiments we carried out. It is possible that with more effort put into engineering this component of the model we would see a better result.

\paragraph{Role of data vs inductive bias}
The other major finding is that whether we use the same or different
data to train the main and auxiliary task has overall little impact:
this is indicated by relatively small differences between
configurations in the {\sc aligned} vs {\sc non-aligned}
condition. The differences that are there tend to favor the {\sc aligned} setting.
This lends supports to the conclusion that the {\sc speech/text} auxiliary task contributes to improved performance on the main task via a strong inductive bias rather than merely via enabling the use of extra data. This is in contrast to many other applications of MTL. 

\paragraph{Impact of parameter sharing design}
The third important effect is about how parameters between the tasks are shared, specifically how the shared and task-specific parts of the speech encoder are apportioned. The configuration with maximum sharing of parameters among the tasks (rows 8 and 13) performs poorly compared to sharing only the lower layers of the encoders for speech and text (i.e.\ rows 6 and 11). Additionally, we see that the inclusion of a text-specific speech encoder {\sc s2t} degrades performance: compare for example row 6 to 7, and row 11 to 12. Thus it is best to have a shared speech encoder whose output is directly used by the {\sc speech/text} task, while the {\sc speech/image} task carries out further transformations of the input via an image-specific speech encoder {\sc s2i}. We can interpret this as the MTL emulating a pipeline architecture to some extent: direct connection of the {\sc speech/text} task to the shared encoder forces it to come up with a representation closely correlated with a written transcription, and then the image-specific speech encoder takes this as input and maps it to a more meaning-related representation, useful for the {\sc  speech/image} task.

In addition to the above patterns of performance on image retrieval we now address our further research questions by investigating selected aspects of the encoder representations.

\paragraph{Speaker invariance} 
Table~\ref{tab:speaker-inv} shows the accuracy of speaker
identification from the activation patterns of the output of encoder
{\sc s2i} for the single task model, the 2-task model, and for the
3-task model which achieved the highest recall@10. The accuracy of the
2 task model is almost three times worse than for the single task model, indicating that the inclusion of {\sc speech/text} strongly drives the learned representations towards speaker invariance. The {\sc text/image} task has only a minor impact.
\begin{table}[htb]
    \centering
    \begin{tabular}{l r}
      Model                 & Accuracy \\\toprule 
      Model 1, {\sc s2i}    & 0.297 \\
      Model 2, {\sc s2i}    & 0.101 \\
      Model 6, {\sc s2i}    & 0.085 \\
    \end{tabular}
    \caption{Speaker identification accuracy for three model configurations. Model numbers refer to rows in Table~\ref{tab:core-results}.}
    \label{tab:speaker-inv}
\end{table}

\paragraph{RSA with regard to textual and visual spaces}
Table~\ref{tab:rsa} shows the RSA scores between the encoder representations of utterances and their representations in the spoken, written and visual modalities. 
\begin{table}[htb]
    \centering
    \begin{tabular}{l r r r}
                  & MFCC & Text      &  Image \\\toprule
Model 1, {\sc s2i} & 0.043     &   0.194   & 0.187 \\
Model 6, {\sc s2i} & 0.030     &   0.212   & 0.222 \\
Model 6, {\sc s2t} & 0.099     &   0.243   & 0.105 \\
Image             & 0.008     &   0.083   & 1.000
    \end{tabular}
    \caption{Pearson correlation between pairwise utterance similarity matrices, for utterances represented by Mean MFCC features, written text, three encoders, and the features of the image corresponding to the utterance. Model numbers refer to rows in Table~\ref{tab:core-results}. Analysis carried out on the single best seed/epoch for each configuration, according to Recall@10.}
    \label{tab:rsa}
\end{table}
Comparing the RSA scores between the {\sc s2i} encoder of model 1 (single task) and model 6 (3 task) we see that the correlations with the textual modality and the visual modality are enhanced while the correlation with the input audio modality drops. This can be interpreted as the {\sc speech/text} task nudging the model to align more closely with the text, which also ends up contributing to the correlation with the image space. For model 6 but using the output of the {\sc s2t} encoder, we see the correlation with the text space is even higher while the correlation with the image space is low. These patterns are what we would expect if {\sc speech/text}  does indeed inject a symbolic inductive bias to the model. Finally, while the RSA score between the textual and visual modalities is low (0.083), nevertheless model 6's encoder {\sc s2i} is moderately correlated with both of these (0.212 and 0.222 respectively). 

\paragraph{Phoneme decoding}
Table~\ref{tab:phoneme-decoding} shows how well phonemes can be
decoded from time-aligned slices of four types of representations:
input MFCC features, the activation patterns of a randomly initialized
{\sc s} encoder, and the activations of the  {\sc s} encoder for two
trained models (1-task and 3-task). Phonemes are most decodable from
the 3-task activation patterns, corroborating that the {\sc
  speech/text} task biases the representations towards a symbolic
encoding of speech.  
\begin{table}[htb]
    \centering
    \begin{tabular}{l r}
      Representation                 & Accuracy \\\toprule
      MFCC                  & 0.284 \\
      Random init, {\sc s}& 0.486 \\
      Model 1, {\sc s}    & 0.528 \\
      Model 6, {\sc s}    & 0.578 \\
    \end{tabular}
    \caption{Phoneme decoding accuracy for the four representations. Model numbers refer to rows in Table~\ref{tab:core-results}.}
    \label{tab:phoneme-decoding}
\end{table}

\section{Conclusion}
We show that the {\sc speech/text} task leads to substantial performance improvements when compared to training the {\sc speech/image} task in isolation. Via controlled experiments and analyses we show evidence that this is due to the role of inductive bias on the learned encoder representations.
\paragraph{Limitations and future work}
Our current model does not include an explicit speech-to-text decoder, which limits the types of analyses we can perform. For one, it makes it infeasible to carry out an apples-to-apples comparison with a pipeline architecture. Going forward we would like to go beyond matching tasks and evaluate the impact of an explicit speech-to-text decoder as an auxiliary task.

We  are also planning to investigate how sensitive our approach is to amount of data for the auxiliary task. This would be especially interesting given that one motivation for a visually-supervised end-to-end approach is the  un-availability of large amounts of transcribed speech in certain circumstances.

\section*{Acknowledgements}
I would like to thank Afra Alishahi, Lieke Gelderloos and Ákos Kádár, as well as several anonymous reviewers,
for helpful comments and discussion about this work.
\bibliography{biblio}
\bibliographystyle{acl_natbib}

\end{document}